\documentclass[11pt]{article}

\usepackage[]{acl}

\usepackage{times}
\usepackage{latexsym}
\usepackage{booktabs}
\usepackage{multirow}
\usepackage{adjustbox}
\usepackage[T1]{fontenc}
\usepackage[utf8]{inputenc}
\usepackage{microtype}
\usepackage{inconsolata}
\usepackage[most]{tcolorbox}
\usepackage{enumitem}
\usepackage[dvipsnames]{xcolor}
\usepackage{pifont}
\usepackage{booktabs}
\usepackage{makecell}
\usepackage{longtable}
\usepackage{array}
\usepackage{tabularx}
\usepackage{float}
\usepackage{placeins}
 \usepackage{tcolorbox}
\tcbuselibrary{breakable,skins}
\usepackage{enumitem}
\newtcolorbox[auto counter, number within=section]{prompt}[3][]{%
  enhanced,
  breakable,
  colback=#2!5!white,
  colframe=#2!75!black,
  title=\textbf{Box \thetcbcounter: #3},
  fontupper=\normalsize\fontfamily{cmss}\selectfont,
  #1
}
\definecolor{cwarmblue}{HTML}{00BDB6}
\definecolor{cslate}{HTML}{546072}
\newtcolorbox{promptbox}[1][]{
  breakable,
  colframe=cslate,    
  colback=white,     
  coltitle=white,            
  title=#1,                  
  rounded corners,           
  boxrule=0.5mm,             
  boxsep=5pt,                
  toptitle=1mm,              
  bottomtitle=1mm,           
  left=10pt,                 
  right=10pt,                
  top=5pt,                   
  bottom=5pt,                
}
    
\newtcolorbox{promptboxcn}[1][]{
  breakable,
  colframe=black!70,    
  colback=white,     
  coltitle=white,            
  title=#1,                  
  rounded corners,           
  boxrule=0.5mm,             
  boxsep=5pt,                
  toptitle=1mm,              
  bottomtitle=1mm,           
  left=10pt,                 
  right=10pt,                
  top=5pt,                   
  bottom=5pt,                
  fonttitle=\bfseries        
}
\newenvironment{labellist}
  {\begin{description}[leftmargin=0pt, labelindent=0pt,
                       style=nextline, itemsep=2pt, parsep=0pt,
                       topsep=3pt, partopsep=0pt,
                       font=\normalfont\ttfamily]}
  {\end{description}}
\usepackage{graphicx}
\usepackage{tikz}
\usepackage{forest}
\usetikzlibrary{trees,positioning,shapes,shadows,arrows.meta}
\usepackage{todonotes} 
\usepackage{csquotes}
\usepackage{subcaption}
\usepackage{cleveref}

\usepackage{hyperref}
\title{Measuring Research Progress and Linguistic Inclusion in Low-Resource Languages}

\author{\normalsize Atnafu Lambebo Tonja$^{1}$,  Tadesse Destaw Belay$^2$,  Jesujoba Oluwadara Alabi$^3$,  Israel Abebe Azime$^3$,\\
\textbf{\normalsize Benjamin Rosman$^4$, Ehsan Shareghi$^1$,} \\
\textbf{\normalsize Thamar Solorio$^5$, Pontus Stenetorp$^1$, Monojit Choudhury$^5$}\\
\\
\footnotesize
$^1$University College London, UK, $^2$Instituto Politécnico Nacional, Mexico, $^3$Saarland University, Germany, \\
\footnotesize 
$^4$University of the Witwatersrand, South Africa, $^5$MBZUAI, UAE\\ }
\begin{document}
\maketitle
\begin{abstract}
Large Language Models (LLMs) have accelerated research on low-resource and multilingual NLP, but their effectiveness in creating an equitable landscape remains unmapped. We analyze 1,158 peer-reviewed open-access papers published between 2020 and 2025, examining research attention to low-resource languages (LRLs) and representation across language and speaker populations. To assess linguistic inclusion beyond publication counts, we introduce the Linguistic Inclusion Framework (LIF), a multidimensional framework that measures linguistic inclusion in four dimensions: research growth, representation expansion, representation continuity, and distributional equity. While annual publications grew from 52 to 415, a 5.4$\times$ increase in their ACL Anthology publication share, this growth largely concentrated on represented languages rather than expanding coverage of underrepresented languages.
Further analysis shows that languages spoken in Asia account for 64.1\% of speakers across LRLs we analyze but receive only 46.7\% of the language representation share (0.7$\times$ their proportional share), while European languages receive 5.3$\times$ their proportional share. At the country level, linguistically diverse regions such as sub-Saharan Africa and Southeast Asia show the lowest language coverage ratios.
Applying the framework at the continental level, we find that no region scores well on all four dimensions, showing that research growth alone is not a reliable metric of progress toward equitable and inclusive language technology\footnote{We will release all collected and annotated data for research purposes}.
\end{abstract}

\section{Introduction} \label{intro}
There are more than 7,000 languages spoken in the world, however only a few languages are represented in the development and use of digital language technology \cite{ kornai2013digital,nekoto-etal-2020-participatory,zaugg2022digitally}. This imbalance is clearly shown in the recent development and deployment of large language models (LLMs) \cite{alabi-etal-2025-charting,khanna2025invisible,pava2025mind}. LLM-based systems are now being integrated into domains such as search engines \cite{kirsten-etal-2026-characterizing}, healthcare chatbots \cite{tao2026llm}, educational platforms \cite{kasneci2023chatgpt}, public services \cite{10.1093/scipol/scag026}, and workplace tools \cite{brynjolfsson2025generative}. As these systems become integral to everyday life, limited language support creates a digital divide between communities, as these models are mostly trained on high-resource languages with large amounts of digital data, and their knowledge and reasoning abilities tend to favor the languages and cultures represented in their training data \cite{mihalcea2025ai}. 

      
Recently, there has been notable progress in multilingual dataset creation, LLM training and adaptation, and benchmarking, alongside growing interest in low-resource languages (LRLs) \cite{adelani2022masakhaner,ahuja2024megaverse,pfeiffer2020mad,tonja-etal-2026-afri,ustun2024aya}. However, more papers, models, datasets, or benchmarks do not necessarily indicate greater linguistic inclusion, as progress may remain concentrated in a small subset of already represented languages. Existing metrics capture research output and model performance, but not whether linguistic representation is becoming more inclusive. Thus, they are insufficient for assessing progress toward linguistic inclusion.

Prior work has examined methodological and regional developments in low-resource NLP and highlighted structural inequalities in language representation\cite{bird-2022-local,hedderich-etal-2021-survey,nigatu-etal-2024-zenos,joshi-etal-2020-state}. Beyond the research literature, longitudinal tracking of global AI adoption shows a similar pattern: usage is increasing in both the Global North and Global South, while the gap between them continues to widen \cite{misra2025measuring}. Yet, we lack longitudinal evidence of whether the rapid expansion of LRL research in the LLM era\footnote{We use \emph{LLM era} to refer to the period from 2020 onward, dated from the release of GPT-3 \citep{brown2020language}.} has translated into broader and sustained linguistic representation. In particular, it remains unclear whether this growth has expanded language representation in model development and evaluation, increased resource availability, or narrowed the gap between research coverage and real-world linguistic diversity. 

To address this gap, we map peer-reviewed open-access research on LRLs published between 2020 and 2025. We analyze publication trends, 
explore the geographic and linguistic distribution of research attention and examine inequality in the global distribution of language-level research attention. We also include longitudinal analysis of publicly accessible datasets and models using HuggingFace\footnote{https://huggingface.co/} for the 20 most retained languages from each content to assess if languages receiving continued research attention also show growth in publicly available resources.
By combining these analyses, we introduce Linguistic Inclusion Framework (LIF) a multidimensional framework to measure progress in LRL research. In this framework, we measure linguistic progress across four dimensions: \textbf{Research growth}, \textbf{Representation expansion}, \textbf{Representation continuity}, and \textbf{Distributional equity} (defined in Sec \ref{LIP}). Our analysis is guided by the following research questions:
    \textbf{RQ1:} How does research on LRL change between 2020 and 2025 in terms of publications, tasks, modalities, and contribution types?
     \textbf{RQ2:} How is research attention to LRL distributed across languages, speaker populations, countries, and continents?
     \textbf{RQ3:} Does the growth of LRL research reflect expansion to new languages and sustained representation, or concentration around the same small set of languages? 
     \textbf{RQ4:} Does continued research attention to a language increase publicly accessible datasets and models?

The contributions of this study are: \textbf{1)} We present a mapping of research on LRL in the LLM era, covering publication trends, research topics, modalities, contribution types and language representation between 2020 and 2025. \textbf{2)} We introduce a language-level representation analysis that separates publication growth from linguistic expansion. \textbf{3)} We measure country-level language coverage and compare research representation with real-world linguistic diversity. \textbf{4)} We provide a temporal analysis of language continuity using four categories to assess whether language inclusion is consistent or occurs only occasionally. 
\textbf{5)} We propose a linguistic inclusion framework for measuring research progress in LRL that can be applied at country, language family or continent level. 

\section{Methodology}
In this section, we describe the data collection and annotation pipeline used to answer the research questions. 





\begin{figure*}[!h]
    \centering
    \scriptsize
    \includegraphics[width=\linewidth]{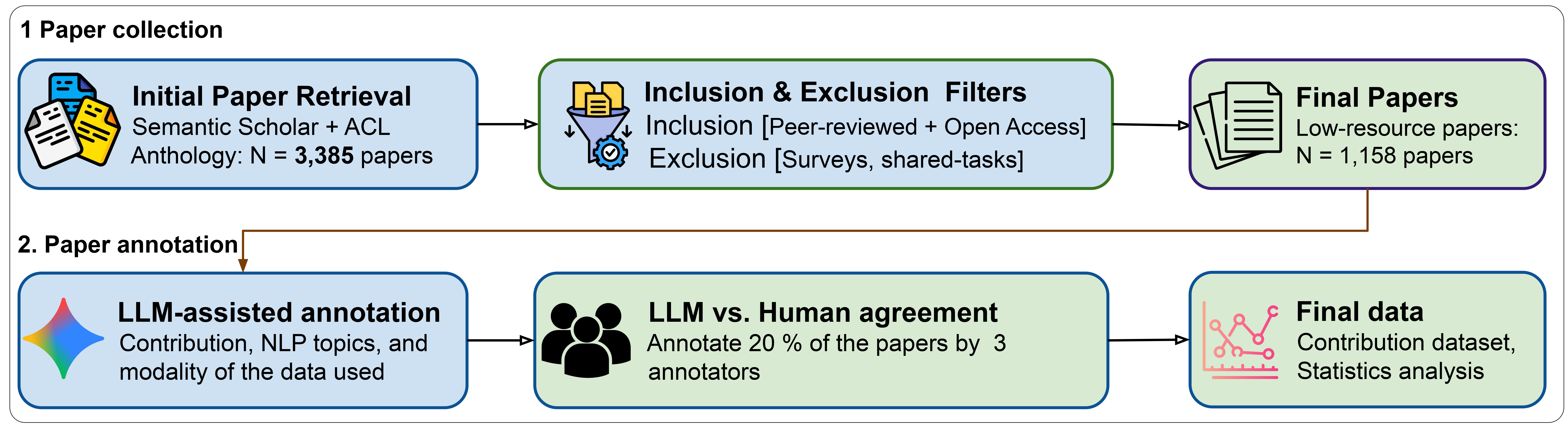}
    \caption{Overall paper retrieval and annotation pipeline.}
    \label{fig:paper-filtering}
    \vspace*{-5mm}
\end{figure*}

\subsection{Data Collection and Filtering} 
\paragraph{Paper collection} We collected papers from Semantic Scholar\footnote{\url{https://www.semanticscholar.org/}} and ACL Anthology.\footnote{\url{https://aclanthology.org/}} We searched for papers published between 2020 and 2025 to assess how research attention to multilingual and LRL has evolved during the LLM era. To filter papers, we created a set of 52 keyword variants containing a combination of the terms "low-resource languages" and "LLMs" and searched for papers containing these terms in their titles or abstracts. 
We added a full list of keywords in Appendix~\ref{sec:appdata}. We merged papers from the two sources and removed duplicates with identical titles. Through this keyword search, we collected 3,385 papers that met our search criteria.


\paragraph{Filtering criteria} We removed non-peer-reviewed papers (e.g., arXiv), surveys, shared tasks, publications on programming languages, and non-open-access publications. This process resulted in a total of 1,403 peer-reviewed open-access papers. Figure~\ref{fig:paper-filtering} provides the complete screening pipeline and the number of papers removed at each stage. 

\paragraph{Research scope classification } \label{scope_filter}
As \citet{nigatu-etal-2024-zenos} mentioned, including the term "\textit{low-resource}" in a paper's abstract or title does not necessarily mean the paper includes or works on LRL. We therefore performed a second screening to select only papers that work on or include LRL in their study, using the LLM-assisted scope classification described in section \ref{llm-pipline}. 
Each paper is assigned to one of the four categories: \textbf{(1)} low-resource language, \textbf{(2)} low-resource domain/task,
\textbf{(3)} simulated low-resource, or
\textbf{(4)} not low-resource language. Appendix Table \ref{tab:low-resource_dist} presents the relative fractions of these types in our filtered list of papers.


\subsection{LLM-assisted Data Annotation} \label{llm-pipline}
We used Gemini 2.5 Pro \cite{gemini25} to extract the covered languages, contributions, NLP topics addressed and modalities involved in each paper. For each paper, we converted the PDF to machine-readable text and provided the LLM with the first 45,000 characters for classification.


\paragraph{Research scope filtering} 
Papers were assigned to one of four categories: \textbf{low-resource language}- referring to papers involving data collection, model development, experimentation, or evaluation in LRL, \textbf{low-resource domain/task}- referring to papers addressing limited data in a specific domain or task while conducting experiments in high-resource languages (HRLs), \textbf{simulation-only}- referring to papers using HRL data to simulate an LRL setting, or \textbf{not low-resource}- referring to papers that did not meet any of these criteria. We follow \citet{joshi-etal-2020-state} in defining LRLs as classes 0 to 3 and HRLs as classes 4 and 5. The full classification criteria, prompt and the number of papers classified in each category are provided in Appendix~\ref{scope_prompt}. We retained 1,158 papers classified as LRL.

\paragraph{Extracting languages} Using a procedure similar to the one described above, we extracted the languages covered in each included paper. We considered a language represented if it appeared in the paper's dataset, model pretraining, adaptation or fine-tuning, benchmark, evaluation, or analysis. We then mapped the identified languages to the countries and continents where they are spoken. For language coverage analyses, we followed the classification of \citet{joshi-etal-2020-state} and retained languages in resource classes 0--3, including varieties and dialectal forms classified independently within these classes. We excluded languages in classes 4--5, we provide detailed prompts used and further details in Appendix~\ref{lang_prompt}.

\paragraph{Paper contribution type} We extracted the paper contribution types and classified them into: \textit{Artifact} and \textit{Knowledge}, comprising eight subcategories following the guidelines and taxonomy proposed in \citet{pramanick-etal-2025-nature}. \textbf{Artifact contributions} includes creating new research resources and are further divided into methods, datasets, and tasks, while \textbf{Knowledge contributions} include works that provide insights that advance understanding of existing NLP components including methods, datasets, tasks, languages, and people. We provide detailed prompts used in Appendix~\ref{contr_prompt}.

\paragraph{NLP topics} We adopted the taxonomy used by the ACL Rolling Review (ARR) submission system,\footnote{\url{https://aclrollingreview.org/cfp}} which includes 29 NLP research topics, and further extracted one or more NLP research topics for each paper that reflect its primary scope using the same model as used before. We provide detailed prompts used in Appendix~\ref{topic_prompt}.


\paragraph{Modality type} Similarly, we extracted primary data modality used in each paper and categorized them into: \textit{text}, \textit{speech}, \textit{image}, \textit{multimodal}, or \textit{other}. We classified papers that integrate multiple modalities as \textit{multimodal}, and studies that did not fit the predefined categories as \textit{other}. We provide detailed prompts used in Appendix~\ref{modality_prompt}.



\subsection{Human Verification}
To assess the accuracy of LLM-assisted data extraction and annotation, we recruited three annotators \footnote{All annotators contributed to the study and are included as co-authors.} with NLP research experience. We randomly sampled 20\% of the LLM-annotated data for verification. For each sample, annotators marked \textbf{Yes} for each field if the LLM-extracted values were correct and \textbf{No} otherwise. We measured pairwise human-human agreement and human-LLM agreement across five annotation tasks: research scope classification, language extraction, contribution type extraction, NLP topic extraction, and modality extraction.
The human annotators showed high pairwise agreement, reaching up to 98.67\%, indicating strong consistency in the annotation process. 
The lowest human-human agreement was observed for low-resource categorization and Language extraction, at 90.26\% and 91.97\%, respectively. More importantly, human-LLM agreement was also high across the five annotation tasks, reaching up to 100\%. The lowest human-LLM agreement was observed for low-resource categorization and language extraction, at 96.68\% and 97.32\%, respectively, indicating that the LLM annotations were generally consistent with human judgments. We provide detailed results in Appendix~\ref{sec:apphumanllm}.


\section{State of Research on LRLs}\label{progress}
Using the annotations, we examine how LRL research progressed from 2020 to 2025. We analyze trends in publication volume, linguistic representation, and resource trajectories.

\subsection{Publication Trends} \label{pub}
\begin{figure}[h!]
    \centering
    \includegraphics[width=\linewidth]{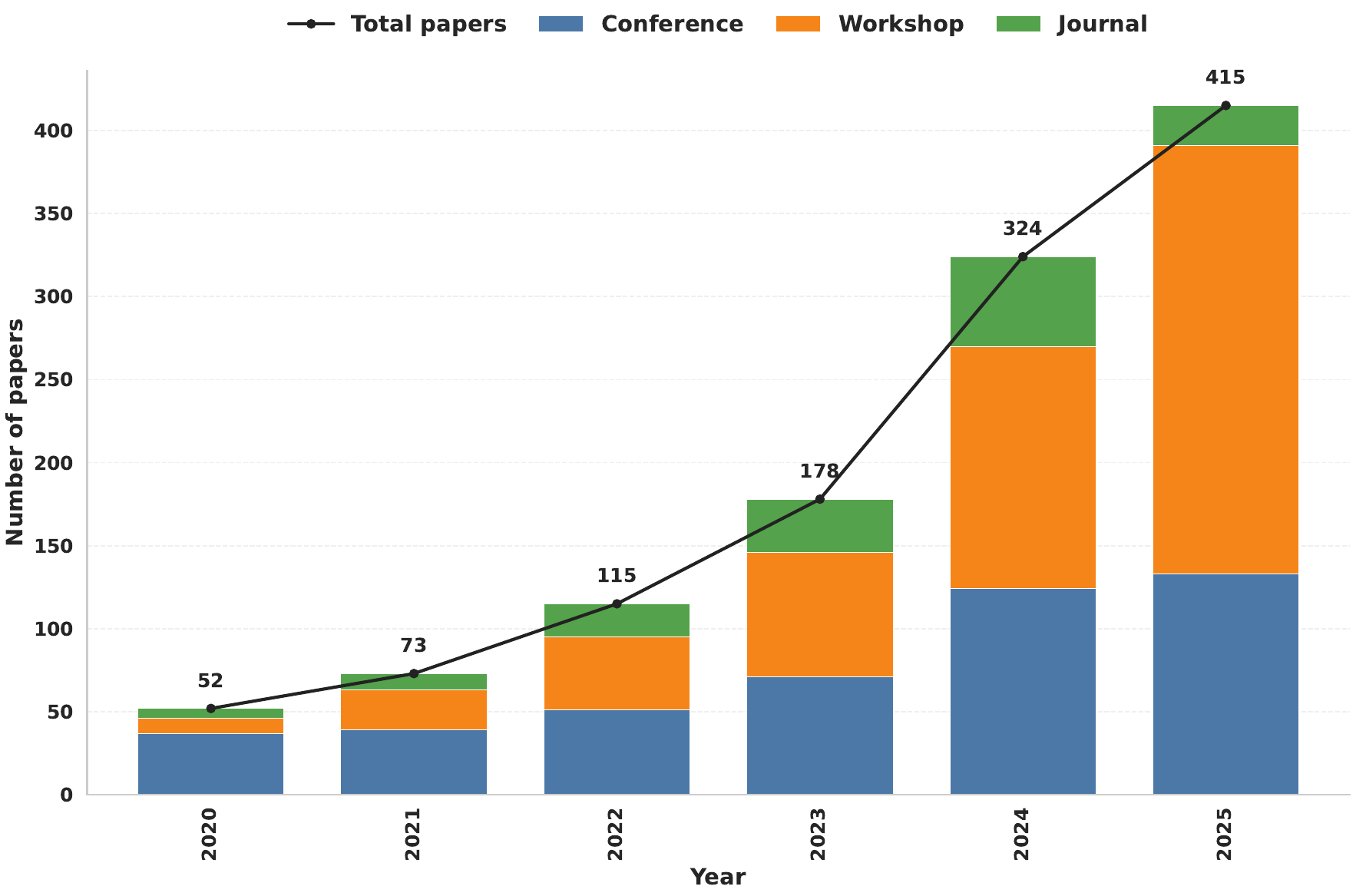}
    \caption{Year-by-year growth in LRL research publications from 2020 to 2025, aggregated by publication venue.}
    \label{fig:pub_trend}
    \vspace*{-5mm}
\end{figure}

Figure \ref{fig:pub_trend} shows year-by-year publication statistics aggregated by venues. We observe that research publications on LRLs have increased during the LLM era, coinciding with the emergence and widespread adoption of LLMs. This trend suggests a growing research attention toward multilingual and low-resource NLP, but does not necessarily imply adequate representation or support of the majority of LRLs or their speaker communities.

Publication count alone cannot separate growth in LRL research from growth in NLP as a whole. To assess whether the increase reflects a shift in NLP research attention or overall expansion, we examined publications in our dataset that appear in ACL Anthology, which account for 70.5\%(816) of our collected papers. Within this subset, LRL papers grew 11$\times$ between 2020 and 2025, while total ACL Anthology publications grew 2.05$\times$, this shows the LRL share of annual Anthology publications increased from 0.43\% to 2.33\% (see Appendix \ref{app:share}). The increase in LRL research therefore reflects a shift in research attention, not only overall research expansion.   

For publication venues, we observe a larger increase in workshop publications over time. The growth in workshop publications may be partly associated with the creation of dedicated venues for LRLs and regional language technologies, such as AfricaNLP, MRL, and AmericasNLP. These venues provide platforms for emerging research communities and help disseminate preliminary findings, datasets, and benchmarks in a fast-evolving field. However, the decline in journal publications in 2025 may reflect longer publication cycles than those of conferences and workshops. 

Furthermore,  we analyzed the authors' institutional affiliations in our corpus and observed that institutions in the USA, Canada, Germany, the UK, China, and India account for the largest share of publications, while those in Africa and South America are comparatively underrepresented, as shown in Figure~\ref{fig:affiliation-world-map}. This suggests that research on LRLs in these regions is often conducted outside these continents, showing the concentration of linguistic research in the Global North. 
\subsection{NLP Topics, Contributions and Modalities}
\begin{figure*}[t]
    \centering
    \scriptsize
    \begin{subfigure}{0.48\textwidth}
        \centering
        \includegraphics[width=\linewidth]{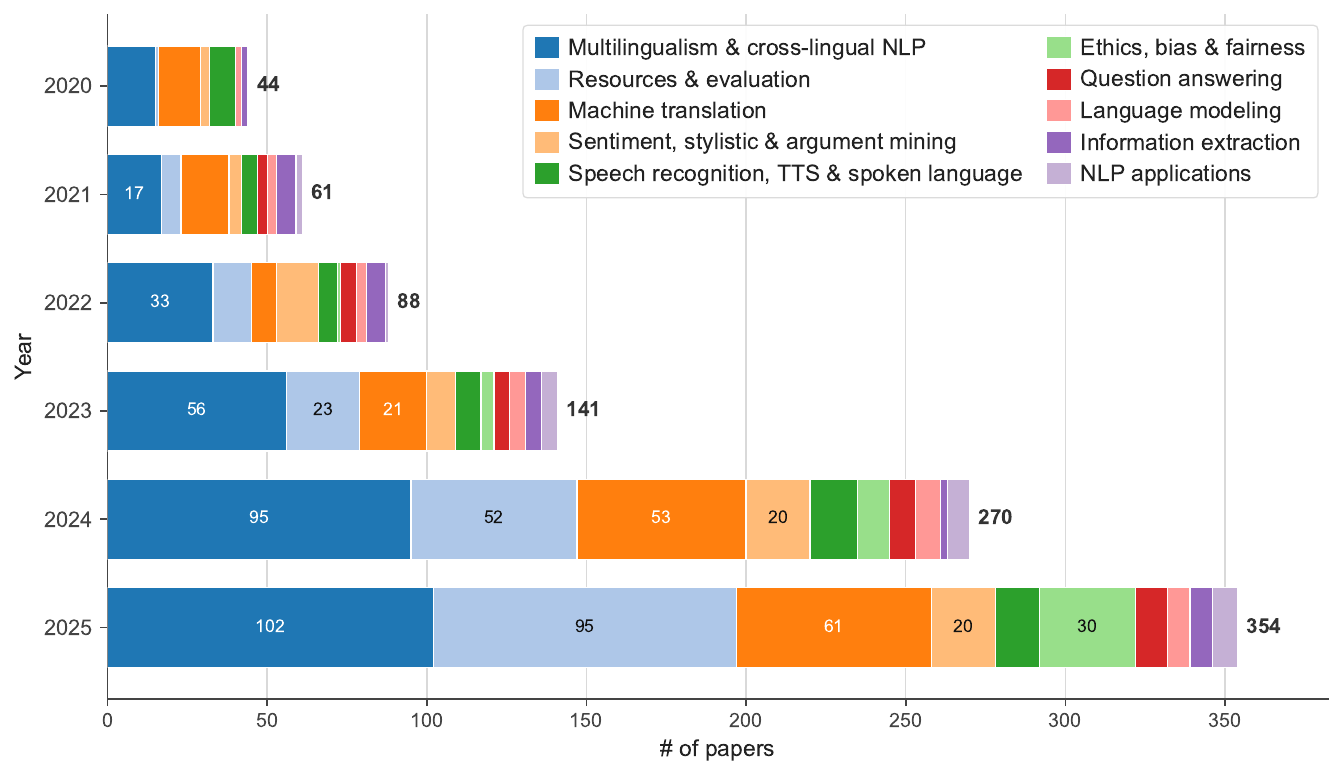}
        \caption{Publication trends}
        \label{fig:nlp-tasks}
    \end{subfigure}
    \hfill
    \begin{subfigure}{0.48\textwidth}
        \centering
        \includegraphics[width=\linewidth]{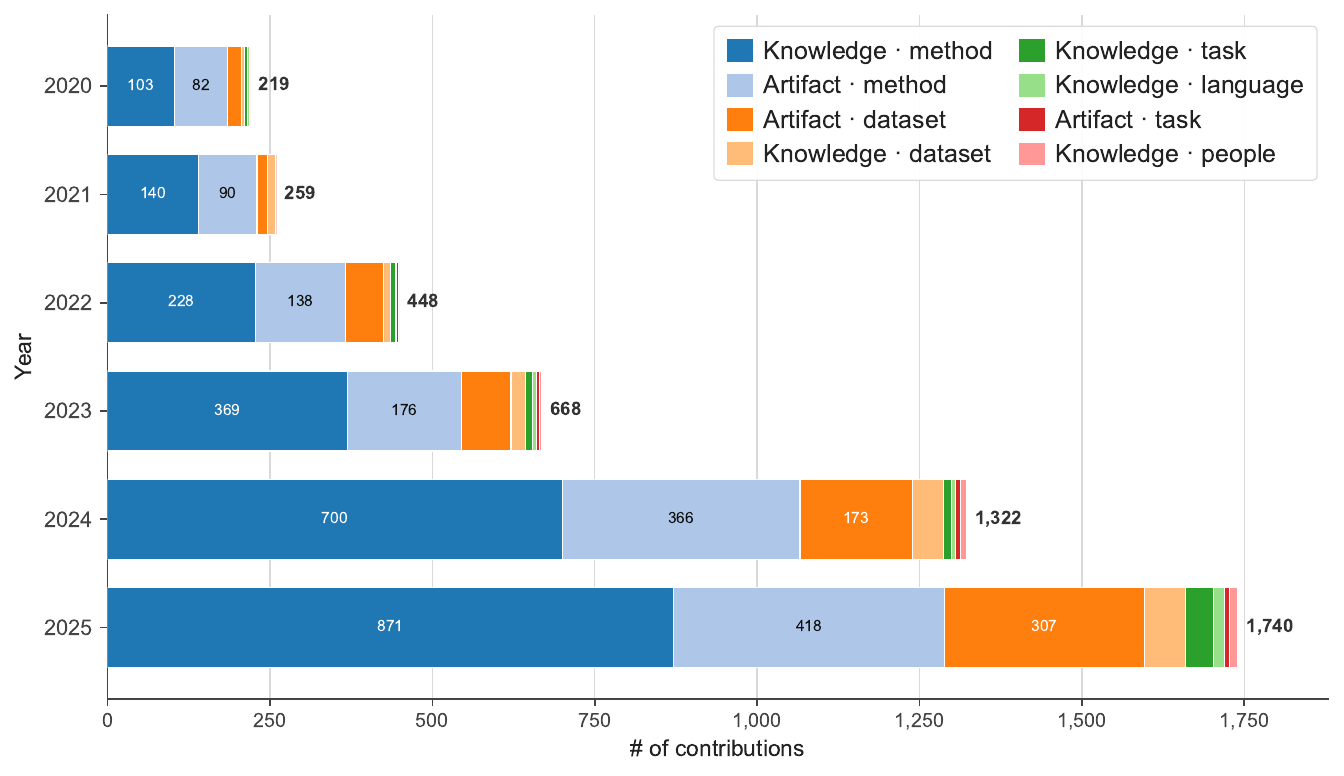}
        \caption{Topic distribution}
        \label{fig:contribution}
    \end{subfigure}

    \caption{Overview of publication trends and research topics.}
    \label{fig:overview}
    \vspace*{-5mm}
\end{figure*}

\paragraph{NLP topics}
Figure \ref{fig:nlp-tasks} shows the distribution of NLP topics across the collected papers. NLP topics provide a comprehensive, standardized characterization of research focus in LRLs and facilitate topic-based analysis, retrieval, and benchmarking. The chart shows that LRL research spans diverse NLP topics, dominated by Multilingual and cross-lingual NLP (24.8\%), Resources and evaluation (23.1\%), and Machine translation (14.8\%).  Together, these three topics account for nearly two-thirds of the published literature.

However, classical NLP tasks such as syntax/parsing, morphology, and information extraction receive relatively little attention.  This pattern suggests a shift in low-resource NLP priorities toward cross-lingual transfer, dataset curation, evaluation, and alignment.

\paragraph{NLP Contributions} 
As shown in Figure \ref{fig:contribution}, the distribution of claim statements follows a consistent pattern over years, with \textit{Knowledge method} accounting for the largest share, followed by \textit{Artifact method} and \textit{Artifact dataset}. In 2025, these categories account for 50.1\%, 24.0\%, and 17.6\%, respectively, together representing 91.7\% of all claim statements. Notably, \textit{Artifact dataset} has consistently increased since 2022. However, conceptual contributions that focus directly on tasks, languages, or social/human dynamics remain marginal. This pattern suggests that contributions in low-resource NLP are predominantly framed around methodological innovations and dataset/artifact, rather than qualitative language-specific or human-centric analysis.\raggedbottom
\paragraph{Types of Modalities}
As shown in Figure \ref{fig:domain} despite the rise of multimodal research, text remains the dominant modality, accounting for over 80\% of papers in 2025. In contrast, other modalities remain underrepresented despite modest growth in 2025, with speech and multimodal research accounting for just 4.9\% and 6.6\% of publications, respectively. This highlights a critical research gap in LRLs across other modalities. 
\paragraph{Takeaway.} To answer \textbf{RQ1}, research on LRLs increased substantially in publication volume between 2020 and 2025, growing 5.4$\times$ as a share of ACL Anthology publications and driven largely by growth in workshop publications. This growth is unevenly distributed, with research concentrated in a limited number of topics. Contributions are predominantly method- and dataset-focused, while text remains the dominant modality, highlighting gaps in research on other modalities for LRLs.



\subsection{Language Coverage} \label{lan_coverage}
In this section, we assess (1) country-level language coverage, (2) speaker-weighted representation for each country, and (3) regional distribution of speaker and language representation. 
\paragraph{A) Country-level language coverage}

To assess LRL coverage per country, we compute the ratio of unique languages represented in our data to the total number of languages spoken in each country. We used Ethnologue\footnote{https://www.ethnologue.com/statistics/} to identify languages spoken in each country. We calculate country-level coverage ratio as follows:
\begin{equation}
\mathrm{LangCoverage}_{c} =\frac{\mathrm{DistinctLang}_{c}}{\mathrm{TotalLang}_{c}} \times 100  
\end{equation}
Where $\mathrm{DistinctLang}_{c}$ is the number of unique languages from country $c$ represented in the papers and $\mathrm{TotalLang}_{c}$ is the total number of languages spoken in that country.  

\begin{figure*}[h!]
    \centering
    \scriptsize
    \begin{subfigure}[b]{0.32\textwidth}
        \centering
        \includegraphics[width=\textwidth]{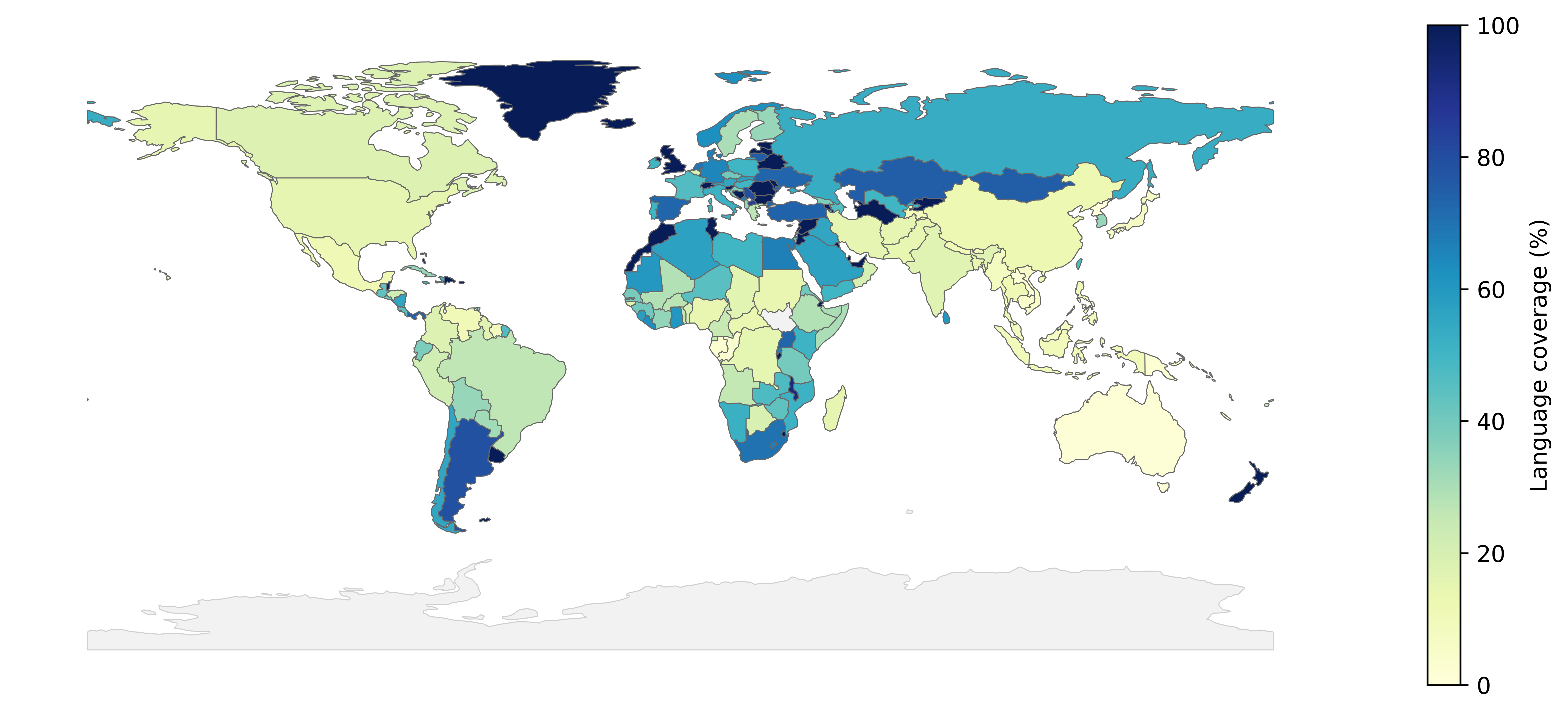}
        \caption{}
        \label{fig:all_map}
    \end{subfigure}
     \hspace{0.01\textwidth}
    \begin{subfigure}[b]{0.32\textwidth}
        \centering
        \includegraphics[width=\textwidth]{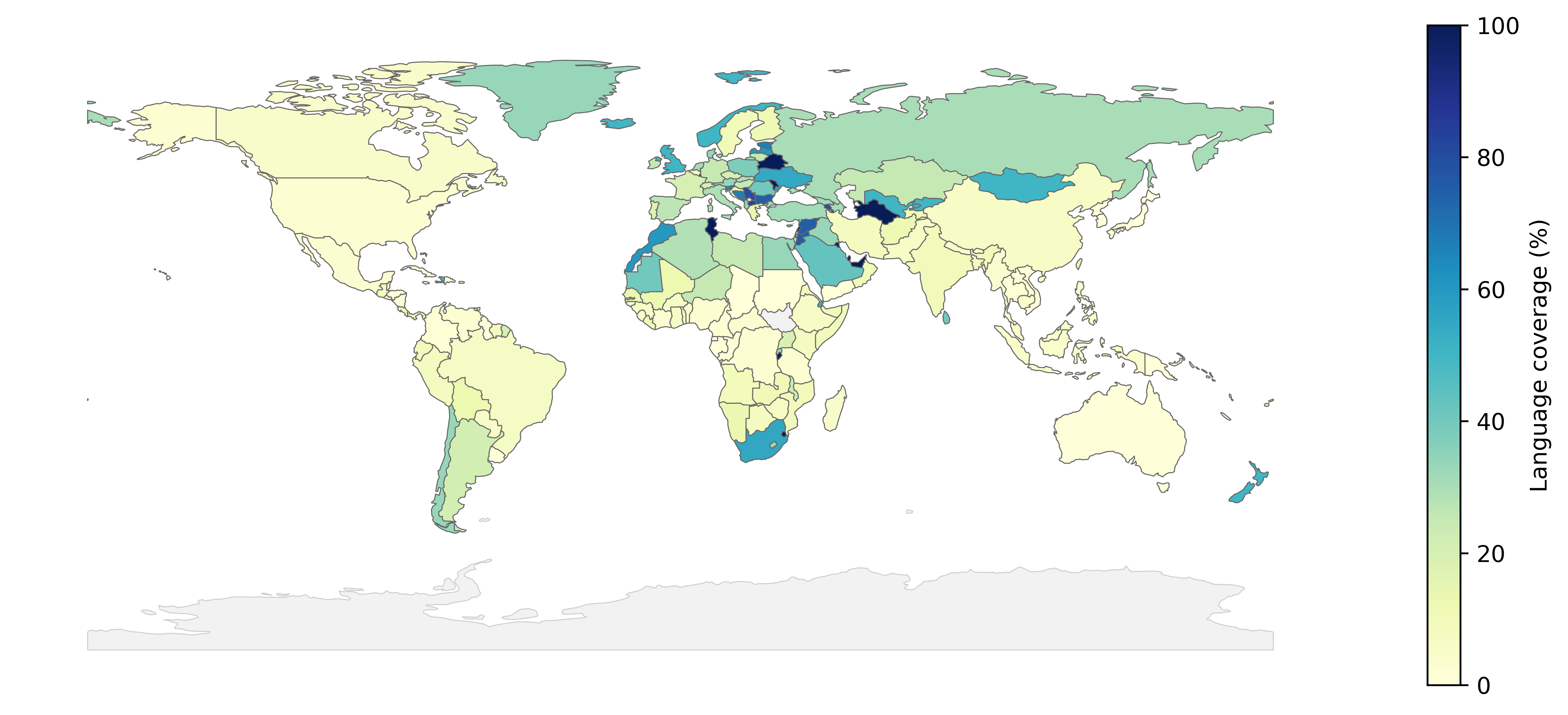}
        \caption{}
        \label{fig:sub_map}
    \end{subfigure}
     \hspace{0.01\textwidth}
    \begin{subfigure}[b]{0.32\textwidth}
        \centering
        \includegraphics[width=\textwidth]{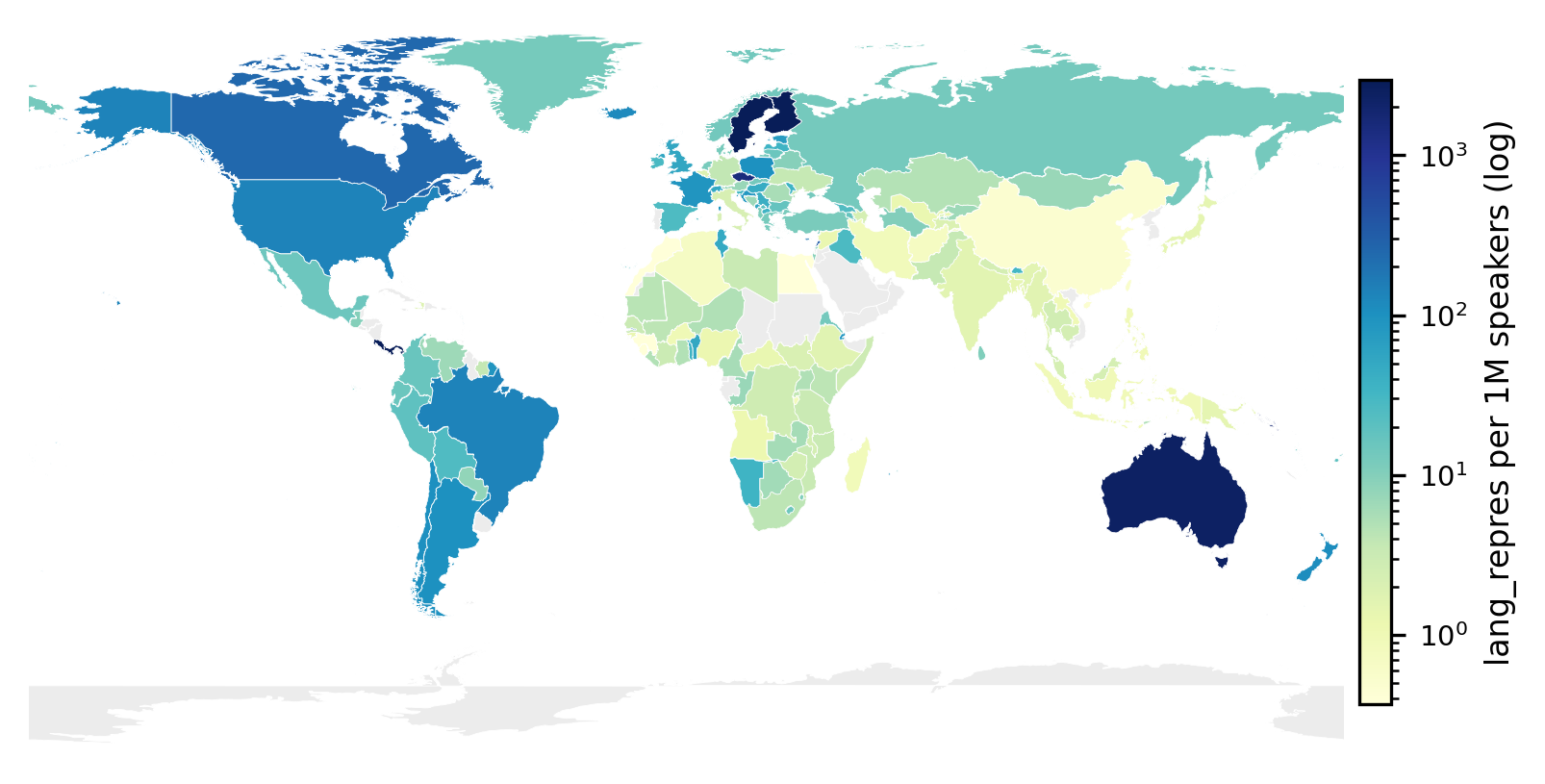}
        \caption{}
        \label{fig:speaker_map}
    \end{subfigure}
    \caption{Country-level distribution of language representation in our dataset. (a) Language coverage ratio across all languages, regardless of the number of papers in which they are represented; (b) language coverage ratio considering only languages represented in more than one paper; and (c) language representation per one million speakers.}
    \label{fig:lang_covarage}
    \vspace*{-5mm}
\end{figure*}

Figure~\ref{fig:lang_covarage} shows the geographical distribution of language coverage. Darker colors indicate countries where a larger proportion of the languages spoken in the country are represented at least once in the included papers, whereas lighter colors indicate countries where most languages are absent. Figure~\ref{fig:lang_covarage}(a) shows relatively high coverage in parts of Europe, North Africa, the Middle East, Central Asia, Southern Africa, southern South America, and New Zealand. In contrast, many linguistically diverse regions remain sparsely represented, indicating that research coverage does not correspond closely to global linguistic diversity. This disparity becomes more pronounced in Figure~\ref{fig:lang_covarage}(b), which considers only languages represented in more than one paper. The substantial reduction in coverage indicates that many languages appear in only a single paper, while repeated research attention concentrates on a smaller set of languages and regions. Thus, although LRLs have received increased attention in the LLM era, this attention remains unevenly distributed across languages and regions.

\paragraph{B) Speaker-weighted representation}
Country-level coverage ratios show the extent to which languages spoken in a country are represented in the papers. However, this measure treats all languages equally, regardless of speaker population size. To analyze how research attention correlates with the number of people it can reach, we weight representation by speaker population size\footnote {We obtained the number of speakers for each language from Ethnologue.}. For each country $c$, we aggregate the languages we extracted from the research papers that are spoken in each country and compute the speaker-weighted representation per one million speakers for each country as follows:
\begin{equation}
    \mathrm{SpeakerRepres}_{c} = \frac{\mathrm{LangRepres}_{c}}{\mathrm{NumofSpeakers}_{c}}\times 10^{6}
\end{equation}

Where $LangRepres_{c}$ is the total number of papers that include the language spoken in country $c$, and $NumofSpeakers_{c}$ is the combined speaker population (L1 \& L2). 
Figure \ref{fig:speaker_map} shows the country-level speaker-weighted attention distribution across continents. As shown in the figure, countries with small speaker populations, e.g., Australia and Iceland, show more than 100 mentions per million speakers, whereas countries with a large number of speakers, e.g India, with 40 represented languages and 1,553 language mentions in the research papers, reach only 1.6 mentions per million speakers. Similarly, Indonesia shows 0.95, Nigeria 1.22, and Morocco 0.20 mentions per million speakers. 

\begin{figure*}[h!]
    \centering
    \scriptsize
    \includegraphics[width=\textwidth]{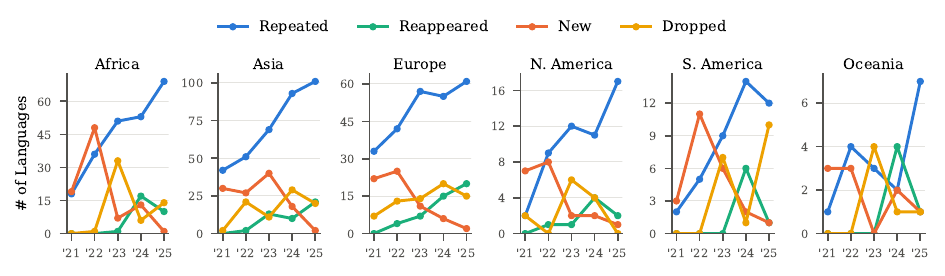}
    \caption{Year-to-year language transition by categories and continent.} 
    \label{fig:transition}
    \vspace*{-5mm}
\end{figure*}

We provide further details in Appendix \ref{sec:appcontient}. Our analysis also shows that languages from Asia and Africa are underrepresented relative to their share of speakers, whereas languages from other regions are comparatively overrepresented.

\paragraph{Takeaway} These findings answer \textbf{RQ2}: research attention to LRLs is unevenly distributed across languages, speaker populations, countries, countries and continents.

\subsection{Language Expansion and Continuity}\label{langexpansion}

To assess whether increased language representation in research reflects broader linguistic inclusion or repeated attention to the same set of languages, we analyze year-to-year transitions in language representation. 
We classify each transition into four categories: \textbf{New} - if the language appears for the first time in the later year, \textbf{Repeated} - if it appears in both consecutive years, \textbf{Reappeared} - if the language returns after being absent in the previous year, having been present at some point before and \textbf{Dropped}- if the language is present in the earlier year but not in the following year. 



Figure \ref{fig:transition} shows the distribution of transition categories by continent. Across continents, year-to-year transitions show that repeated languages dominate in all regions, while the proportions of new, reappeared, and dropped languages vary by continent.  
When we look at new-language inclusion over time across continents, it is consistently smaller than repeated-language inclusion. Across all continents, repeated languages represent 57.2\% of all transition instances compared to only 19.6\% for newly included languages. On average, we observed an increase in new language inclusion in the 2020 $\rightarrow$ 2021 transition, with new language representation of 43.5\%. However, in 2024$\rightarrow$2025, new language inclusion decreased to 2.1\%, while repeated language representation increased to 68.5\%. This suggests the increase in research activity reflects repeated attention to a selected set of languages rather than consistent inclusion of previously underrepresented languages. Similarly, languages that drop and reappear suggest that some languages appear in one year but are not represented the following year, and others receive attention in later years after a period of absence, with dropped languages representing 14.7\% and reappeared languages representing 8.5\% of all transition instances. 
Research growth in LRL therefore reflects sustained attention to a selected set of languages rather than consistent inclusion of previously underrepresented languages. 

\paragraph{Takeaway} These findings answer \textbf{RQ3}: research growth in the LLM era reflects concentration around a small set of already represented languages rather than sustained expansion to underrepresented languages.    
%

\paragraph{Resource availability} Our analysis showed that repeated languages consistently dominate over the inclusion of new languages across all continents. We next ask whether this sustained attention to a small set of languages translates into increase in resource availability for those languages. We analyze the growth in datasets and models released on Hugging Face for the top 20 most frequently studied languages across six geographical regions between 2020 and 2025. We identify language-specific resources using the Hugging Face API and count each resource based on its initial appearance on the Hub. Multilingual resources are counted for each target language they cover. Details on resource identification are provided in Appendix~\ref{sec:hf-collection}.

As shown in Figure \ref{fig:language-class-aggregates}, these resources have generally increased across the six regions, with particularly notable growth among Asian languages. In contrast, languages from North and South America have experienced comparatively limited growth in the number of datasets and models released during this period. This provides a partial answer to \textbf{RQ4}: among the most frequently repeated languages, publicly available resources grew over the study period; however, this growth is not uniformly distributed across regions and remains concentrated in languages that were already well represented.

\begin{figure}[h!]
    \centering
    \includegraphics[width=\linewidth]{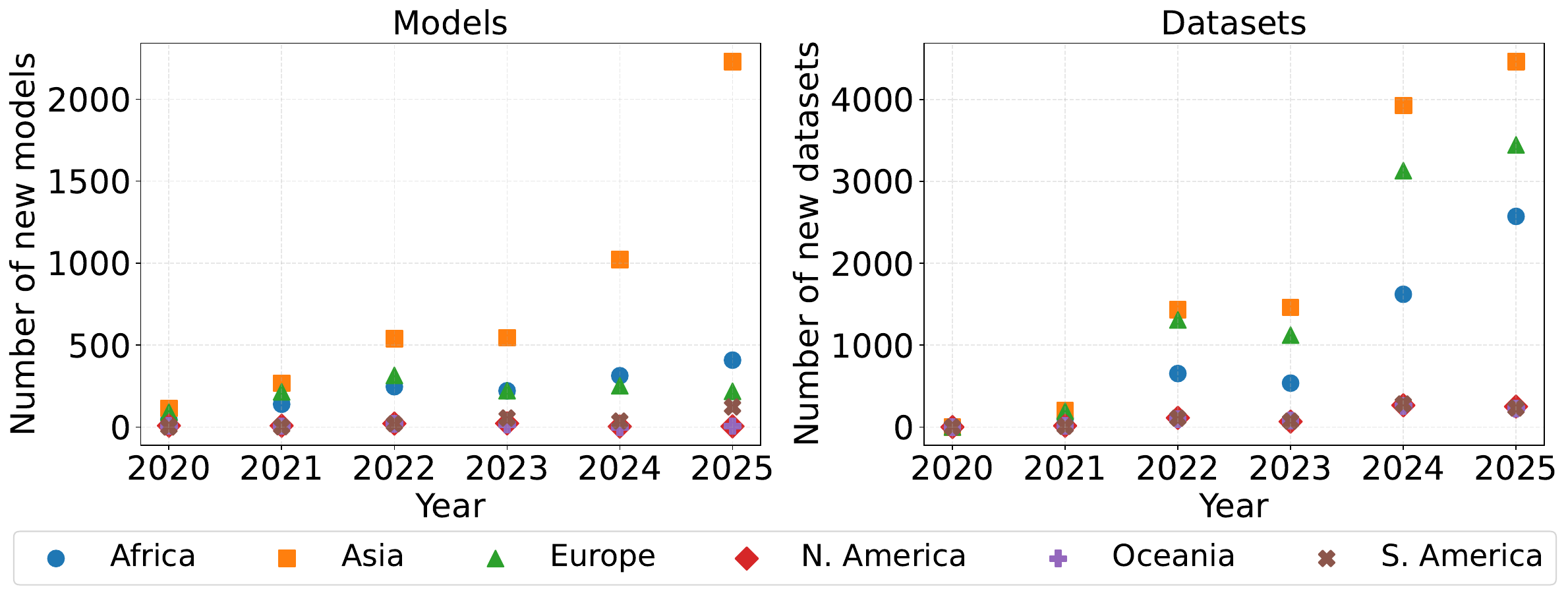}
    \caption{Year-by-year growth in newly released HuggingFace models (left) and datasets (right).}
    \label{fig:language-class-aggregates}
    \vspace*{-5mm}
\end{figure}

\begin{table*}[!h]
\centering
\scriptsize
\setlength{\tabcolsep}{4pt}
\renewcommand{\arraystretch}{1.3}
\begin{tabular}{p{1.9cm}p{4.1cm}p{4.8cm}p{4cm}}
\toprule
\textbf{Dimension} & \textbf{What it shows} & \textbf{Measurement} &
\textbf{A higher score indicates} \\
\midrule
Research growth &
Whether research activity involving LRLs increases over time &
$\dfrac{\text{least-squares slope of annual instances}}{\text{mean annual instances}}$ &
Stronger growth relative to the average research activity \\
Representation expansion &
Whether research extends beyond the languages represented at the start of the study &
$\dfrac{\text{languages first seen after 2020}}{\text{all languages, 2020--2025}}$ &
More new languages entered the literature after the baseline year \\
Representation continuity &
Whether represented languages remain visible rather than appearing once &
$\dfrac{\text{repeated}}{\text{repeated} + \text{dropped}}$ & More sustained representation, a lower score means languages disappear after first appearing \\
Distributional equity &
Whether attention is evenly shared among LRLs or concentrated in a subset &
$1 - \text{Gini}$ of instances per language, within continent &
A more even distribution \\
\bottomrule
\end{tabular}
\caption{The four dimensions of the LIF. \emph{Instances} are language-publication instances,
and \emph{repeated} and \emph{dropped} are counted over all year-to-year
transitions.}
\label{lip}
\vspace*{-5mm}
\end{table*}

\section{Measuring Linguistic Inclusion} \label{LIP}

In Section \ref{progress}, we have seen 
increase in the number of publications and language-publication instances during the study period for LRLs. However, country-level language coverage is not uniformly distributed across regions
and research attention is concentrated on a small set of already represented languages. Based on the analysis, no framework measures these dimensions together.    
To address this, we propose a \textbf{Linguistic Inclusion Framework (LIF)}, a multidimensional framework to measure progress in LRL research and its linguistic inclusion
across four dimensions: \textit{research growth}, \textit{representation expansion}, \textit{representation continuity}, and \textit{distributional equity}. In Table \ref{lip}, we define each dimension and how we calculate it. The proposed framework can be used for any set of languages with per-year publications.

\subsection{Framework Application and Discussion}
We illustrate the LIF using our data at the continental level from 2020-2025 period, the same computation applies at any level of grouping. 
\begin{table}[t]
\centering
\small
\resizebox{\columnwidth}{!}{%
\begin{tabular}{@{}lrrrrrrr@{}}
\toprule
\textbf{Continent}  &
\textbf{Langs}  & \textbf{Inst.} & \textbf{Growth} &
\textbf{Expan.} & \textbf{Cont.} & \textbf{1$-$G} & \textbf{Top-10} \\
\midrule
Africa  & 97&1{,}856 & 0.348 & 0.814 & 0.810 & 0.335 & 50.5\% \\
Asia &138 &3{,}520 & 0.328 & 0.674 & 0.816 & 0.311 & \textbf{43.0\%} \\
Europe  &91& 1{,}713 & 0.264 & 0.571 & \textbf{0.821} & 0.391 & 45.5\% \\
North America  &27& 248 & 0.319 & 0.852 & 0.785 & 0.464 & 75.8\% \\
South America  &22& 186 & 0.306 & \textbf{0.909} & 0.692 & 0.404 & 84.4\% \\
Oceania &10 & 65 & \textbf{0.382} & 0.900 & 0.739 & \textbf{0.605} & 100.0\% \\\hline
\end{tabular}}
\caption{LIF results per continent, for dimensions defined in Table \ref{lip}. 
We added top-10 shares because equity is inflated where the represented languages are few.}
\label{lipresult}
\vspace{-3mm}
\end{table}
Table \ref{lipresult} shows the results for four dimensions when applying our data at the continent level. We see positive growth in research across all continents, however, the relative increases differ. Continents such as Oceania and Africa show the greatest growth in research, while Europe shows the weakest relative increase. For new language expansion, South America and Oceania show the highest growth, while Africa and North America also show substantial growth. Europe and Asia have lower new language expansion scores, indicating that a larger share of their represented languages were already represented at the beginning of 2020. 

For language retention, Europe has the strongest language retention, followed by Asia and Africa. This indicates that languages represented in research from these regions are comparatively more likely to remain visible across consecutive years. South America has the lowest language retention despite having the highest new language inclusion. This suggests that the newly included languages are not retained. We computed distributional equity using only represented languages therefore, its result depends on the number of represented languages for each continent. Oceania shows the highest score with ten languages, and Asia shows the lowest with 138 languages. We also report the top-10 share, the proportion of a continent's language-publication instances received by its ten most-represented languages; lower values indicate broader distribution.  Oceania has a 100\% top-10 share, represented by ten languages. Attention is most widely distributed in Asia (43\% top-10 share), followed by Europe (with a 45.5\% top-10 share). This indicates that research attention is distributed across a wider set of languages in regions where more languages are represented.     


No continent shows high scores across all four dimensions. Even in the continent where concentration is lowest, attention remains unequally distributed. Africa's ten most-represented languages account for 50.5\% of its instances across 97 represented languages, and Asia's top 10 account for 43\% across 138 languages. Asia shows low expansion but better continuity. Europe combines strong continuity with weakest growth while South America shows the most new language inclusion but lowest continuity. This suggests each inclusion dimension requires different interventions. In conclusion, these results show that research growth and expansion into new languages do not by themselves produce sustained or equitable representation across languages. We hope that the proposed framework will encourage future assessments of LRL research to move beyond publication counts and model performance to more comprehensive assessments of language inclusion.   

\section{Conclusion}
We mapped 1,158 peer-reviewed open-access papers on LRs published between 2020 and 2025 to assess whether the growth of research in the LLM era has translated into broader linguistic inclusion. We saw growth in the number of publications however, this growth was concentrated in a limited set of already represented languages. We also found that research attention does not align with both linguistic diversity and speaker populations. Year-to-year transitions show that repeated languages increasingly dominate while new language inclusion declines. To measure progress and linguistic inclusion in LRL research, we propose the LIF, which measures progress across four dimensions. We found that no region performs well on all four dimensions when applying LIF to our data at the continental level. 

\section{Limitation}
Our data is restricted to peer-reviewed, open-access papers retrieved from Semantic Scholar and the ACL Anthology using English keywords. It therefore under-represents works published in other languages and in closed-access venues. Because our search terms combine low-resource and LLM-related keywords, part of the observed growth may reflect changing terminology rather than research activity alone. Language, contribution, topic, and modality annotations were produced by an LLM. We rely on \citet{joshi-etal-2020-state} classes and Ethnologue for language classification, country mapping, and speaker counts. These sources carry their own gaps and uncertainties, especially for dialects, sign languages, and languages spoken across borders. Language-paper counts treat a mention in a massively multilingual benchmark the same as a paper dedicated to a single language, which may overstate attention to some languages. 
\bibliography{custom}
\clearpage
\raggedbottom
\appendix

 \section{Related Work}
 

Several surveys, including~\citet{hedderich-etal-2021-survey}, have examined NLP for low-resource languages from different perspectives. General surveys cover methodological advances such as data augmentation~\citep{chen-etal-2023-empirical}, neural machine translation~\citep{Ranathunga2021NeuralMT,haddow-etal-2022-survey}, application~\citep{Mbaye2025OpportunitiesAC}, and pre-trained (large) LMs~\citep{unjum2025llms,Hussen2025TheSO,lupacscu2026large}. In addition, a growing number of surveys focus on specific languages or language families, including African~\citep{tonja-etal-2023-natural,Issaka2024TheGN,InuwaDutse2025NaijaNLPAS,alabi-etal-2025-charting, Belay2025TheRO}, Latin American~\citep{tonja-etal-2024-nlp}, Asian~\citep{liu-best-2025-survey}, and other regional languages, highlighting their unique linguistic characteristics, available resources, and application challenges.

Complementing these surveys are several position papers focused on low-resource NLP. Existing position papers have separately addressed digital decay~\citep{kornai2013digital}, deployment constraints ~\citep{joshi-etal-2019-unsung}, structural inequalities in language representation~\citep{joshi-etal-2020-state}, and the conceptual instability of the term "low-resource"~\citep{bird-2022-local,nigatu-etal-2024-zenos}. In parallel, recent work proposes a shift toward dynamic language acquisition and open-ended discovery paradigms~\citep{f-p-dossou-aidasso-2025-towards}. Recent empirical work such as \citet{Nwatu2026AtlasNLPAC} further highlights persistent geographic and country-level imbalances in the representation of NLP datasets.

Despite this growing body of work, existing surveys and position papers leave several gaps. First, many surveys focus on a specific NLP task or methodological area, such as applications, data augmentation, machine translation, or language modeling. Second, surveys focused on particular languages or language families provide detailed regional perspectives but do not offer a unified view of low-resource NLP as a global research area. Third, several of the existing broad surveys predate the rapid emergence of LLMs, leaving the recent transformation of low-resource NLP underexplored. In particular, there is limited systematic synthesis of how the LRLs landscape has changed in the LLM era, including shifts in research focus, datasets and resources, model development and evaluation, language and geographic coverage, and emerging challenges. Building on both the survey and position paper literature, this study provides a unified and up-to-date synthesis of LRL research, with particular attention to how the field has evolved in the LLM era.

\section{Data Collection and Keywords}
\label{sec:appdata}
To gather data we used search queries by combining terms describing language models with terms used in the literature to describe languages with limited resources or representation. Each phrase was enclosed in quotation marks to enable exact phrase matching. The general query structure we used is as follows:

\begin{quote}
\centering
\texttt{"language-model term" "language-resource term"}
\end{quote}

Every language-model term in the first row of Table~\ref{tab:core-search-terms} was combined with every language-resource term in the second row. This procedure produced 35 core search queries.

\begin{table}[htbp]
\centering
\scriptsize
\begin{tabularx}{\columnwidth}{
    >{\raggedright\arraybackslash}p{0.22\columnwidth}
    >{\raggedright\arraybackslash}X
}
\toprule
\textbf{Term group} & \textbf{Exact search terms} \\
\midrule

Language-model terms
&
\texttt{"large language model"};
\texttt{"LLM"};
\texttt{"pre-trained language model"};
\texttt{"foundation model"}; and
\texttt{"generative AI"}. \\
\addlinespace

Language-resource terms
&
\texttt{"low-resource language"};
\texttt{"low-resourced language"};
\texttt{"under-resourced language"};
\texttt{"underrepresented language"};
\texttt{"endangered language"};
\texttt{"minority language"}; and
\texttt{"indigenous language"}.
\\

\bottomrule
\end{tabularx}
\caption{Terms used to construct the core literature-search queries.}
\label{tab:core-search-terms}
\end{table}

We also used the supplementary queries shown in Table~\ref{tab:supplementary-search-terms}. These queries captured alternative descriptions of language-resource scarcity and more specific references to
GPT and multilingual models.

\begin{table}[htbp]
\centering
\scriptsize
\begin{tabularx}{\columnwidth}{
    >{\raggedright\arraybackslash}p{0.24\columnwidth}
    >{\raggedright\arraybackslash}X
}
\toprule
\textbf{Query group} & \textbf{Exact queries} \\
\midrule

GPT
&
\texttt{"GPT" "low-resource language"}; 
\texttt{"GPT" "low-resource"}; 
\texttt{"GPT" "underrepresented language"}; and
\texttt{"GPT" "minority language"}.
\\[0.6em]

Alternative resource terminology
&
\texttt{"large language model" "less-resourced language"}; 
\texttt{"large language model" "resource-scarce language"}; 
\texttt{"large language model" "low-density language"}; 
\texttt{"large language model" "data-scarce language"}; 
\texttt{"large language model" "heritage language"}; 
\texttt{"large language model" "underserved language"}; and
\texttt{"large language model" "marginalized language"}.
\\[0.6em]

Additional LLM terminology
&
\texttt{"LLM" "less commonly taught language"}; 
\texttt{"LLM" "resource-poor language"}; and
\texttt{"LLM" "less-studied language"}.
\\[0.6em]

Multilingual models
&
\texttt{"multilingual language model" "low-resource"}; 
\texttt{"multilingual model" "under-resourced"}; and
\texttt{"multilingual NLP" "low-resource language"}.
\\

\bottomrule
\end{tabularx}
\caption{Supplementary literature-search queries.}
\label{tab:supplementary-search-terms}
\end{table}

\section{Affiliation}
\begin{figure}[h!]
  \centering
  \includegraphics[width=\linewidth]{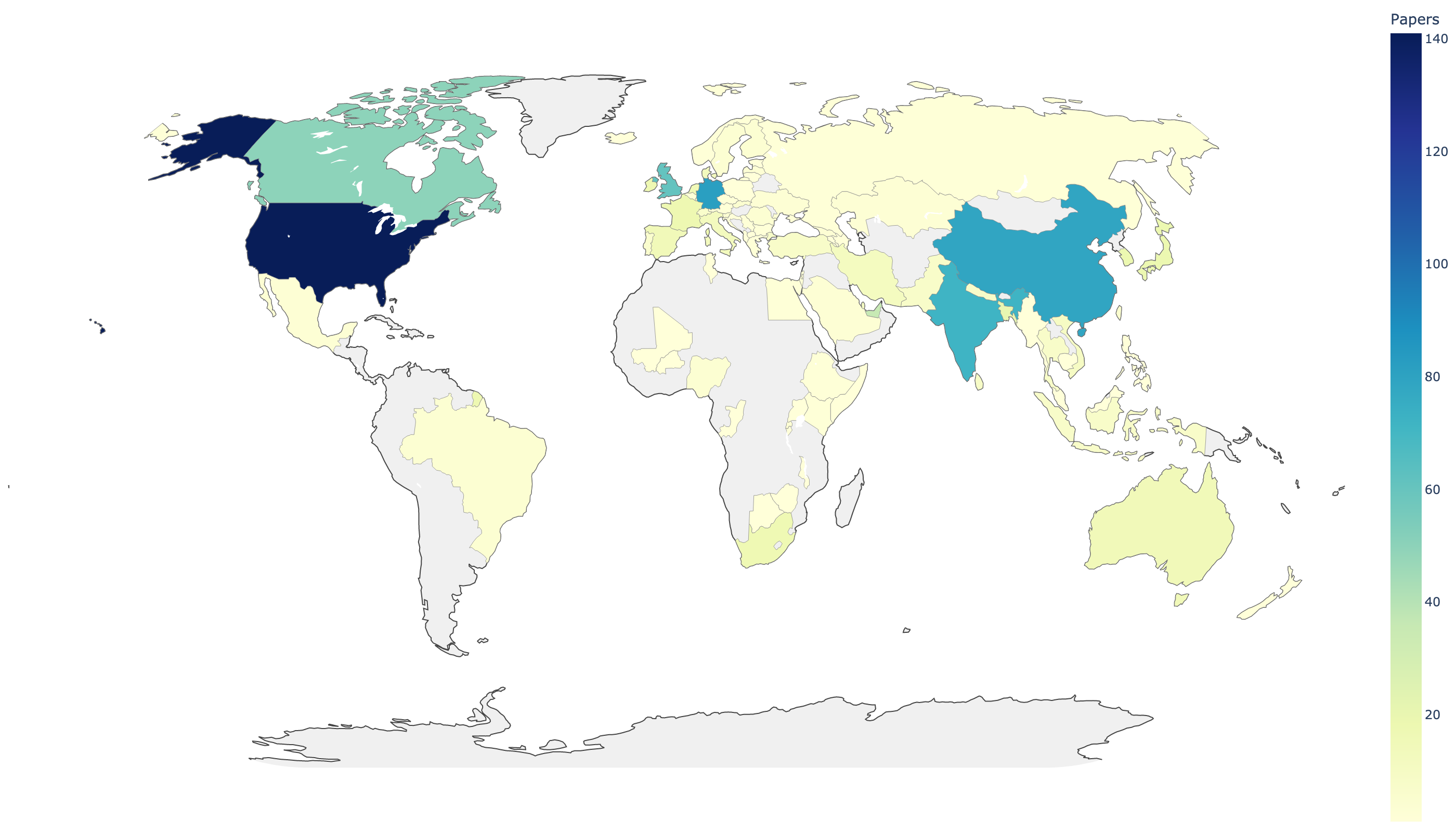}
  \caption{Geographic distribution of low-resource NLP papers by the
  countries of \emph{all} author affiliations. Each paper is counted once
  per distinct country appearing among its affiliations. Darker shades
  indicate more papers; grey denotes countries with no mapped
  affiliations. The United States (135), Germany (82), China (77),
  India (72), and the United Kingdom (61) contribute the most papers.}
  \label{fig:affiliation-world-map}
\end{figure}
\FloatBarrier

\section{LLM-Assisted Annotation} \label{sec:appllmannot}
\subsection{Scope filtering}\label{scope_prompt}
The following prompt was used to classify each paper according to its
relationship with low-resource languages and settings. The paper text supplied
to the model was truncated to the first 45,000 characters.

\begin{promptboxcn}[Prompt used for low-resource category verification]
{\small
You are an expert in natural language processing (NLP), multilingual NLP, and
low-resource language research.

Analyze the research paper text and assign the paper to \textbf{exactly one}
of the following categories:

\begin{enumerate}
    \item \textbf{low-resource languages}: The study meaningfully includes at
    least one genuinely low-resource language in its dataset, model
    pretraining, adaptation or fine-tuning, benchmark, evaluation, or
    experimental analysis. This category applies even if the study includes
    many high-resource languages. Languages mentioned only in the
    introduction, motivation, related work, examples, or references do not
    count.

    Examples of high-resource languages include English, Spanish, German,
    Japanese, French, Arabic, Mandarin Chinese, Swedish, Turkish, Korean,
    Hindi, Persian, Portuguese, Czech, Russian, Dutch, Polish, Croatian,
    Italian, Vietnamese, Basque, Hungarian, Finnish, Serbian, and Catalan.
    Treat languages outside this set as low-resource unless the paper provides
    evidence otherwise.

    \item \textbf{low-resource domain}: The study focuses on a specialised
    domain with limited training or annotated data, such as clinical, legal,
    or historical text. Domain-specific data scarcity, rather than language,
    is the primary bottleneck.

    \item \textbf{simulation only}: The study discusses low-resource settings
    but uses only high-resource languages and artificially creates data
    scarcity through subsampling, truncation, few-shot settings, or another
    form of data restriction.

    \item \textbf{not low-resource}: The study uses standard high-resource
    languages and domains without a low-resource motivation or restricted-data
    experimental setting.
\end{enumerate}

Apply the categories in the order given. In particular, the meaningful
inclusion of at least one low-resource language takes precedence over the
other categories.

\medskip
\textbf{OUTPUT FORMAT}

Return only a JSON object with exactly two keys:

\begin{itemize}
    \item \texttt{"reasoning"}: one short sentence identifying the relevant
    languages, domain, and experimental setting.
    \item \texttt{"category"}: exactly one of
    \texttt{"low-resource languages"},
    \texttt{"low-resource domain"},
    \texttt{"simulation only"}, or
    \texttt{"not low-resource"}.
\end{itemize}

Do not include Markdown, code fences, or text outside the JSON object.

\medskip
\textbf{PAPER TEXT}

\texttt{[Full paper text, truncated to the first 45,000 characters]}
}
\end{promptboxcn}
After applying the scope-classification prompt, we report the number and percentage of papers assigned to each category in Table~\ref{tab:low-resource_dist}.
\begin{table}[htbp]
    \centering
    \begin{tabular}{lrr}
        \toprule
        \textbf{Category} & \textbf{\# Papers} & \textbf{\%} \\
        \midrule
        Low-resource languages & 1,158 & 82.54 \\
        Low-resource domain    &   104 &  7.41 \\
        Simulation only        &    93 &  6.63 \\
        Not low-resource       &    48 &  3.42 \\
        \bottomrule
    \end{tabular}
    \caption{Distribution of papers across the scope-classification categories.}
    \label{tab:low-resource_dist}
\end{table}

\subsection{Language extraction}\label{lang_prompt}
We used the following prompt to extract the languages represented in each research paper. We considered a language represented if it appeared in the paper's dataset, model pretraining, adaptation or fine-tuning, benchmark, evaluation, or experimental analysis.
\begin{promptboxcn}[Prompt used for extracting languages represented in research papers]
{\small
Analyze the research-paper text and extract all languages represented in the research.

\medskip
\textbf{DEFINITION}

A language is considered represented if it is included in the paper's dataset, model pretraining, adaptation or fine-tuning, benchmark, evaluation, or experimental analysis.

\medskip
\textbf{RULES}

\begin{itemize}
    \item Extract every language explicitly included in at least one of the research components specified above.
    \item Include both high-resource and low-resource languages.
    \item Include source, target, training, validation, and test languages when they are part of the research.
    \item Do not include languages that appear only in the background, related work, motivation, illustrative examples, or references.
    \item Do not provide explanations, conversational text, or additional commentary.
    \item If no represented language can be extracted, return \texttt{None}.
\end{itemize}

\medskip
\textbf{OUTPUT FORMAT}

Return the extracted languages as a single comma-separated list:

\begin{quote}
\texttt{Language 1, Language 2, Language 3}
\end{quote}

For example:

\begin{quote}
\texttt{Vietnamese, English, French}
\end{quote}

If no language is represented in the research, return:

\begin{quote}
\texttt{None}
\end{quote}

\medskip
\textbf{PAPER TEXT}

\texttt{[Full paper text, truncated to the first 45,000 characters]}
}
\end{promptboxcn}

\subsection{LLM prompt for NLP Paper Contribution Annotations} \label{contr_prompt}
\begin{promptboxcn}[LLM prompt for annotating abstracts for contribution and NLP tasks]
{\small
You are an expert at analyzing NLP research papers. Based on the title and abstract below, extract the following information:\\
\\
For each sentence in the abstract that describes a contribution, classify it into the appropriate category and extract the exact sentence. Group sentences by contribution type:\\
**artifact-method**: Introduces or proposes a new or novel NLP methodological approach or model to solve NLP tasks. [\textit{example here}]\\
**artifact-dataset**: Introduces a new NLP dataset (i.e., textual resources such as corpora or lexicons). [\textit{example here}]  \\
**artifact-task**: Introduces or proposes a new or novel NLP task formulation (i.e., well-defined NLP problem). [\textit{example here}]\\
**knowledge-method**: Describes new knowledge or insights about NLP models or methods. [\textit{example here}] \\
**knowledge-dataset**: Describes new knowledge about datasets, such as their new properties or characteristics. [\textit{example here}]\\
**knowledge-task**: Knowledge/insights about existing tasks or problem domains. [\textit{example here}]\\
**knowledge-language**: Presents new knowledge about language, such as a new property or characteristic of language. [\textit{example here}]\\
**knowledge-people**: Presents new knowledge about people, humankind, society, or communities. [\textit{example here}]\\
\\
OUTPUT FORMAT:\\
\\
\{\{
  "contribution": \{\{ "artifact-task": ["sentence 1", "sentence 2"],"artifact-method": ["sentence 1"], "artifact-dataset": [], "knowledge-method": ["sentence 1", "sentence 2"],"knowledge-dataset": ["sentence 1"], "knowledge-task": [], "knowledge-language": [], "knowledge-people": [] \}\},\\
  "nlp-tasks": ["task1", "task2", ...]\\
\}\}
\\
INSTRUCTIONS:\\
- A sentence might be assigned to more than one contribution category\\
- Multiple sentences can belong to the same category\\
Title: \{title\}\\
Abstract: \{abstract\}"""\\
Abstract: [PAPER-ABSTRACT]\\
Answer:\}
}
\end{promptboxcn}

\subsection{LLM prompt to Annotate NLP Topics}\label{topic_prompt}
\begin{promptboxcn}[LLM prompt for annotating NLP topics]
{\small
You are an expert at analyzing NLP research papers. Based on the title and abstract below, extract the following information:\\
NLP TOPICS\\
Identify the specific NLP topic (category) addressed from the following list:\\
- \textit{\# List of NLP TOPICS goes here ...}
\\
Output format:\\
\{
  "nlp-tasks": ["task1", "task2", ...]
\}
\\
Title: \{title\}\\
Abstract: \{abstract\}"""\\
Title: [PAPER-TITLE]\\
Abstract: [PAPER-ABSTRACT]\\
Answer:
}
\end{promptboxcn}

\subsection{LLM prompt to annotate modality}\label{modality_prompt}
\begin{promptboxcn}[LLM prompt to annotate  modality]
{\small
Identify the Modality of the data the paper used from the following list:\\
Modality labels\\
Identify the specific modality of the paper from the following list:\\
- \textit{"text", "speech", "image", "multimodal", `"others" }
\\
Output format:\\
\{
  "paper-modality": [modality]
\}
\\
Title: \{title\}\\
Abstract: \{abstract\}"""\\
Title: [PAPER-TITLE]\\
Abstract: [PAPER-ABSTRACT]\\
Answer:
}
\end{promptboxcn}

\section{Human vs. LLM agreement analysis}
\label{sec:apphumanllm}
As shown in Table~\ref{tab:agreement}, pairwise human agreement ranges from 86.02\% to 100\%, while human-vs.-LLM agreement ranges from 96.68\% to 100\%. 
NLP topic categorization has comparatively lower pairwise agreement (86.02\% to 93.33\%), likely because topics overlap and papers may address multiple tasks. For example, named entity recognition may be classified as both information extraction and low-resource NLP, depending on whether the annotation prioritizes the primary task or the broader research setting. 
Pairwise human agreement for LRL categorization ranges from 86.66\% to 96.01\%, reflecting the context-dependent nature of its definition. We follow \citet{joshi-etal-2020-state} in defining a language as a "low-resource language". However, we also classify a language as low-resource if the authors of the paper state that it is low-resource, and we assign it as it is regardless of the categorization used by \citet{joshi-etal-2020-state}. Agreement for language, contribution, and modality extraction ranges from 91.9\% to 100\% between human annotators and from 97.32\% to 100\% between humans and LLMs, with perfect agreement for modality extraction.

\begin{table}[h]
    \centering
    \scriptsize
    \begin{tabular}{l|c|c|c|c}
       \toprule
            \textbf{Agreement} &
            \textbf{Language(s)} &
            \makecell{\textbf{Low-resource}\\\textbf{categorization}} &
            \textbf{Contribution} &
            \textbf{NLP topic} \\
            \midrule
        H1 \& H2 &93.3 &96.01&98.66&92.4\\
        H1 \& H3 &91.97 &90.26&98.66&93.33 \\
        H2 \& H3 &96.49 &91.17&98.67&93.01\\
        Human  \& LLM &97.32 &96.68&99.10&99.11 \\ \bottomrule
    \end{tabular}
    \caption{Pairwise percentage agreement between 
    H1--H3 and the majority-vote human and LLM-generated annotations across the five annotation dimensions.}
    \label{tab:agreement}
\end{table}

\subsection{Annotation guidelines}
\label{app:guidelines}
 
The following text is the guideline document we shared with annotators.
 
\begin{promptboxcn}[Annotation guidelines]
 {
You receive one spreadsheet, one row per paper, which includes paper ID,
title, abstract, venue, year, a link to the PDF, and the model's output
for each task. Each task has an LLM output, a verification, and a
correction column.
 
\begin{labellist}
\item[LLM output] Contains the LLM annotation/extraction output for each
  paper.
\item[Verification] A Yes/No dropdown. Choose \emph{Yes} only when the
  model's output is correct and complete. Anything wrong, anything
  missing, anything extra is \emph{No}.
\item[Correction] Leave empty when you choose \emph{Yes}. When you
  choose \emph{No}, write the correct answer.
\end{labellist}
 
\medskip
\textbf{Task 1: Languages}
 
A language is represented by a paper if (1) it is included in a dataset
the paper collects, curates, annotates, releases or uses; (2) it is in
the data used for pretraining, continued pretraining, adaptation,
fine-tuning or instruction-tuning; (3) it is in a benchmark or test set
the paper evaluates on; (4) it is the source or target of a system the
paper builds, such as translation, ASR, transliteration or OCR; or
(5) it is the subject of an analysis carried out on data.
 
\emph{What does not count.} A language named only in the introduction,
background, related work or motivation; in an illustrative example not
drawn from the paper's own data; in a reference title, or a dataset the
paper names but does not use; in the authors' affiliations,
acknowledgments, or statements about their own languages; in future
work; or in a summary of what other work has covered.
 
\medskip
\textbf{Task 2: Low-resource category}
 
Verify whether the LLM output is correct, and classify each paper into
one of the following.
 
\begin{labellist}
\item[Low-resource languages] The paper runs experiments on, collects
  data for, evaluates on, or builds methods or models for genuinely
  low-resource languages.
\item[Low-resource domain/task] The scarcity is in a specialized
  domain, not a language. The subject matter is the bottleneck.
\item[Simulation-only] The paper discusses low-resource scenarios, but
  every experiment uses high-resource languages, usually by
  artificially restricting the data.
\item[Not-low resource] No meaningful relationship to data scarcity:
  standard languages, standard domains, no scarcity motivation.
\end{labellist}
 
For this study, the following count as high-resource: English, Spanish,
German, Japanese, French, Arabic, Mandarin, Swedish, Turkish, Korean,
Hindi, Persian, Portuguese, Czech, Russian, Dutch, Polish, Croatian,
Italian, Vietnamese, Basque, Hungarian, Finnish, Serbian, Catalan. The
list is based on \citet{joshi-etal-2020-state} resource classes 4 and 5. Do
not add your own judgment to it.
 
\emph{Meaningful inclusion.} A massively multilingual paper counts as
low-resource languages when its low-resource languages are actually in
the work: in the training data, in the evaluation, or in a released
resource. It does not count when they appear only in a description of
some pre-existing model's coverage while every experiment in the paper
is on high-resource languages.
 
\emph{Author self-description.} If an author treats one of the listed
languages as low-resource, use the authors' category as-is.
 
\medskip
\textbf{Task 3: Contribution types}
 
Following the taxonomy of \citet{pramanick-etal-2025-nature}, read the paper and
verify whether the LLM output is correct. Each contribution label is
explained below.
 
\begin{labellist}
\item[artifact-method] Introduces a new method, approach, architecture,
  or model for an NLP task.
\item[artifact-dataset] Introduces a new textual resource: corpus,
  lexicon, treebank, benchmark.
\item[artifact-task] Introduces a new task formulation, a well-defined
  problem not previously posed.
\item[knowledge-method] Reports new insight about models or methods.
\item[knowledge-dataset] Reports new properties or characteristics of a
  dataset.
\item[knowledge-task] Reports insight about an existing task or problem
  domain.
\item[knowledge-language] Reports a property or characteristic of a
  language itself.
\item[knowledge-people] Reports something about people, communities,
  society or the field.
\end{labellist}
 
\medskip
\textbf{Task 4: NLP topics}
 
Verify whether the LLM output matches one of the following NLP topics.
 
\begingroup\raggedright
Phonology, morphology and word segmentation;
Syntax: tagging, chunking and parsing;
Semantics: lexical and sentence-level;
Discourse and pragmatics;
Language modeling;
Natural language understanding;
Question answering;
Information extraction;
Information retrieval and text mining;
Sentiment analysis, stylistic analysis, and argument mining;
Reasoning;
Text generation;
Text summarization;
Machine translation;
Error correction;
Paraphrasing and text rewriting;
Speech recognition, text-to-speech and spoken language understanding;
Sign language processing;
Dialogue and interactive systems;
Conversational AI and chatbots;
Multimodality and language grounding to vision, robotics and beyond;
Vision-language tasks;
Embodied language understanding;
NLP applications;
Computational social science and cultural analytics;
Legal and biomedical NLP;
Educational NLP;
Creative writing and storytelling;
Multilinguality and language diversity;
Dialect and language identification;
Code-switching and multilingual processing;
Low-resource language processing;
Machine learning for NLP;
Representation learning;
Efficient/low-resource methods for NLP;
Transfer learning and domain adaptation;
Few-shot and zero-shot learning;
Resources and evaluation;
Interpretability and analysis of models for NLP;
Benchmarking and testing;
Ethics, bias, and fairness;
Human-centered NLP;
Privacy and security in NLP;
Misinformation detection and fact-checking;
Linguistic theories, cognitive modeling and psycholinguistics;
Computational linguistics theory;
Temporal information processing;
Causal inference in text;
Knowledge graph construction and reasoning;
Text-to-code generation;
Scientific text processing;
Social media and web text analysis;
Document understanding and layout analysis;
Personality and psychological profiling from text;
Others.
\par\endgroup
 
\medskip
\textbf{Task 5: Modality}
 
Verify whether the LLM output correctly describes what the paper
actually processes, based on the following label definitions.
 
\begin{labellist}
\item[text] Written data only: documents, sentences, tokens.
\item[speech] Spoken-language data, alone or paired with text: ASR,
  TTS, speech translation, spoken language understanding.
\item[image] Visual data, alone or paired with text: image
  classification, vision-language tasks, OCR, document images,
  sign-language video.
\item[multimodal] Two or more non-text modalities, or three or more
  modalities in all: speech plus image, audio-visual-text, video with
  audio and subtitles.
\item[others] None of the above.
\end{labellist}
 }
\end{promptboxcn}
\FloatBarrier

\section{LRL papers as a share of ACL Anthology output}
\label{app:share}

\begin{table}[h]
\centering
\begin{tabular}{lrrr}
\toprule
\textbf{Year} & \textbf{Matched} & \textbf{Anthology} & \textbf{Share} \\
\midrule
2020 &  31 &  7{,}163 & 0.43\% \\
2021 &  45 &  6{,}996 & 0.64\% \\
2022 &  75 &  8{,}456 & 0.89\% \\
2023 & 119 &  8{,}936 & 1.33\% \\
2024 & 204 & 11{,}937 & 1.71\% \\
2025 & 342 & 14{,}686 & 2.33\% \\
\midrule
Total & 816 & 58{,}174 & 1.40\% \\
\bottomrule
\end{tabular}
\caption{Papers from our corpus indexed in the ACL Anthology, against total
Anthology output per year.}
\label{tab:share}
\end{table}
\FloatBarrier

\section{NLP Topics, Contributions and Modalities }





\begin{figure}[h!]
    \centering
    \includegraphics[width=0.7\linewidth]{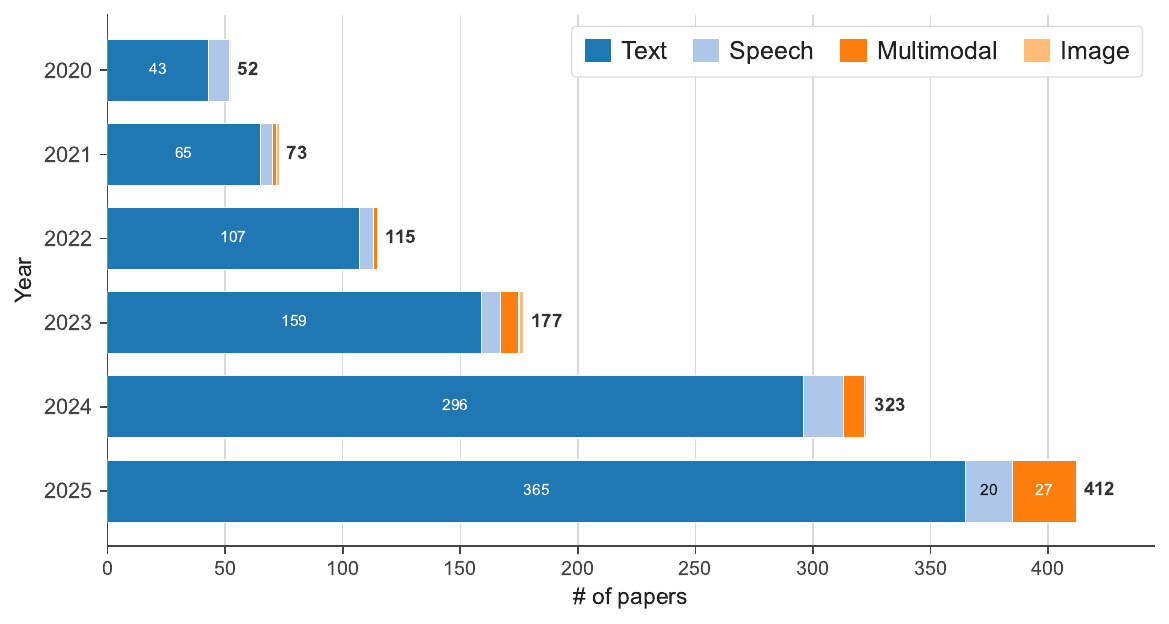}
    \caption{Composition of data modalities by publication year.}
    \label{fig:domain}
\end{figure}
\FloatBarrier

\section{More Results}
\subsection{Continental distribution of speakers and language representation} 
\label{sec:appcontient}
To assess the continental share and its language representation distribution, we compare the continental share of speakers and language representation in our dataset. We calculated the share of speakers and language representation as follows:  
\begin{align*}
\mathrm{Share_{Speaker}}_c &=
\frac{\sum_{l \in c}\mathrm{Speak}_l}
{\sum_{c' \in \mathcal{C}}\sum_{l \in c'}\mathrm{Speak}_l}\times100,\\
\mathrm{Share_{LangRep}}_c &=
\frac{\sum_{l \in c}\mathrm{Rep}_l}
{\sum_{c' \in \mathcal{C}}\sum_{l \in c'}\mathrm{Rep}_l}\times100.
\end{align*}
Here, $\mathcal{C}$ is the set of continents, $l$ denotes a language, $\mathrm{Speak}_l$ is the combined L1 and L2 speaker population of language $l$ reported by Ethnologue, and $\mathrm{Rep}_l$ is the number of analyzed papers in which $l$ is represented. Both measures consider only languages in resource classes 0--3. This allows us to analyze whether the languages spoken in a given region are represented in the literature in proportion to those from other regions.  



\begin{figure}[h!]
    \centering
    \includegraphics[width=0.7\linewidth]{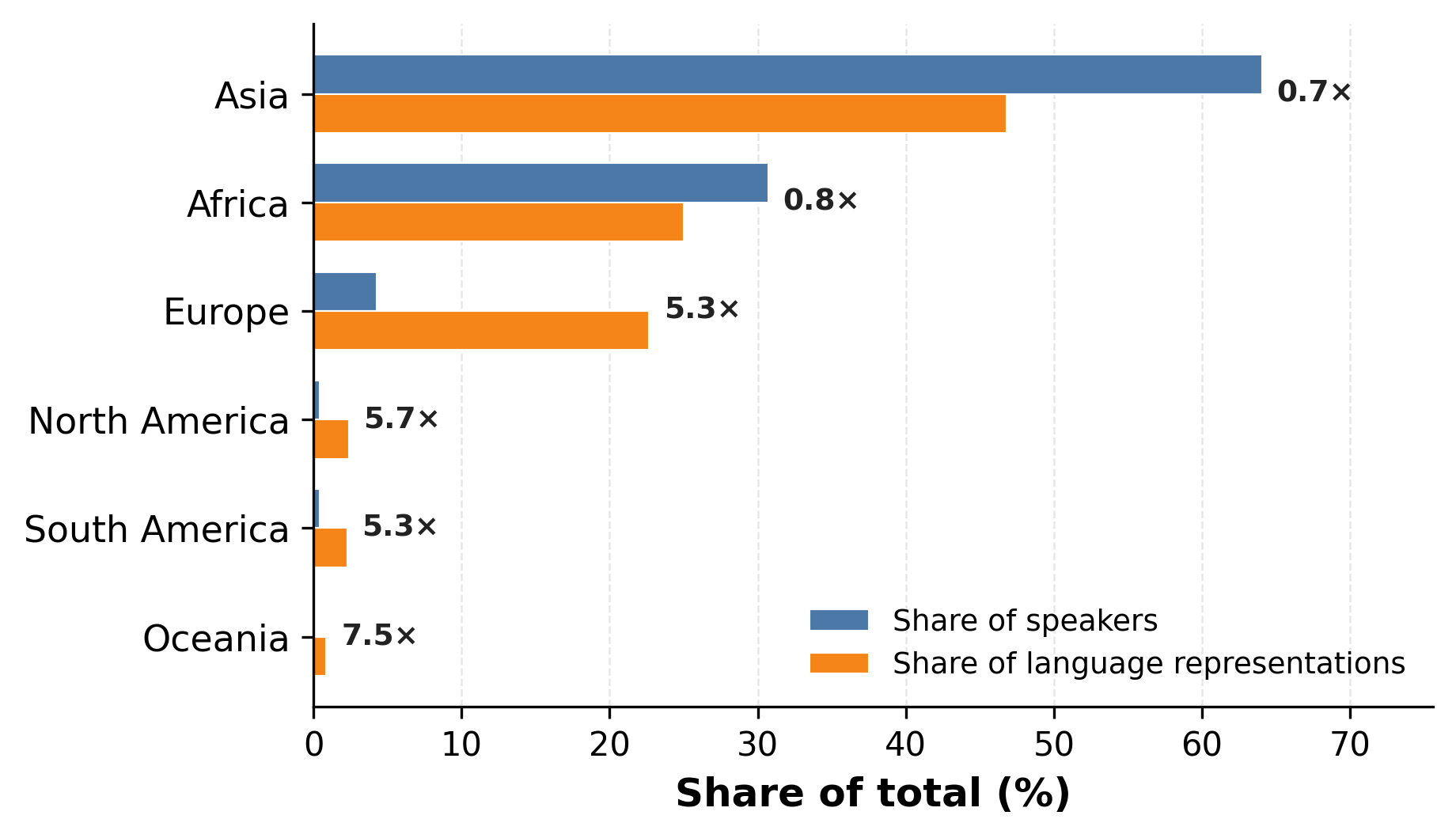}
    \caption{Continental shares of L1\&L2 speakers and language representation.}
    \label{fig:content_shares}
\end{figure}

Figure \ref{fig:content_shares} compares the share of language speakers in each continent with the corresponding share of language representation in the research literature. We observe substantial differences between speaker population and research representation across continents. Languages spoken in Asia account for 64.1\% of the total speaker share but only 46.7\% of language representation, corresponding to 0.7$\times$ their proportional share. Similarly, Africa accounts for 30.7\% of the speaker share but only 24.9\% of language representation (0.8$\times$ its proportional share). However, languages spoken in Europe and other regions receive substantially greater representation relative to their respective speaker shares. Overall, the results show that languages in Asia and Africa are underrepresented relative to their speaker share, while languages in other regions are comparatively overrepresented.

\begin{figure}[h!]
    \centering
    \includegraphics[width=0.7\linewidth]{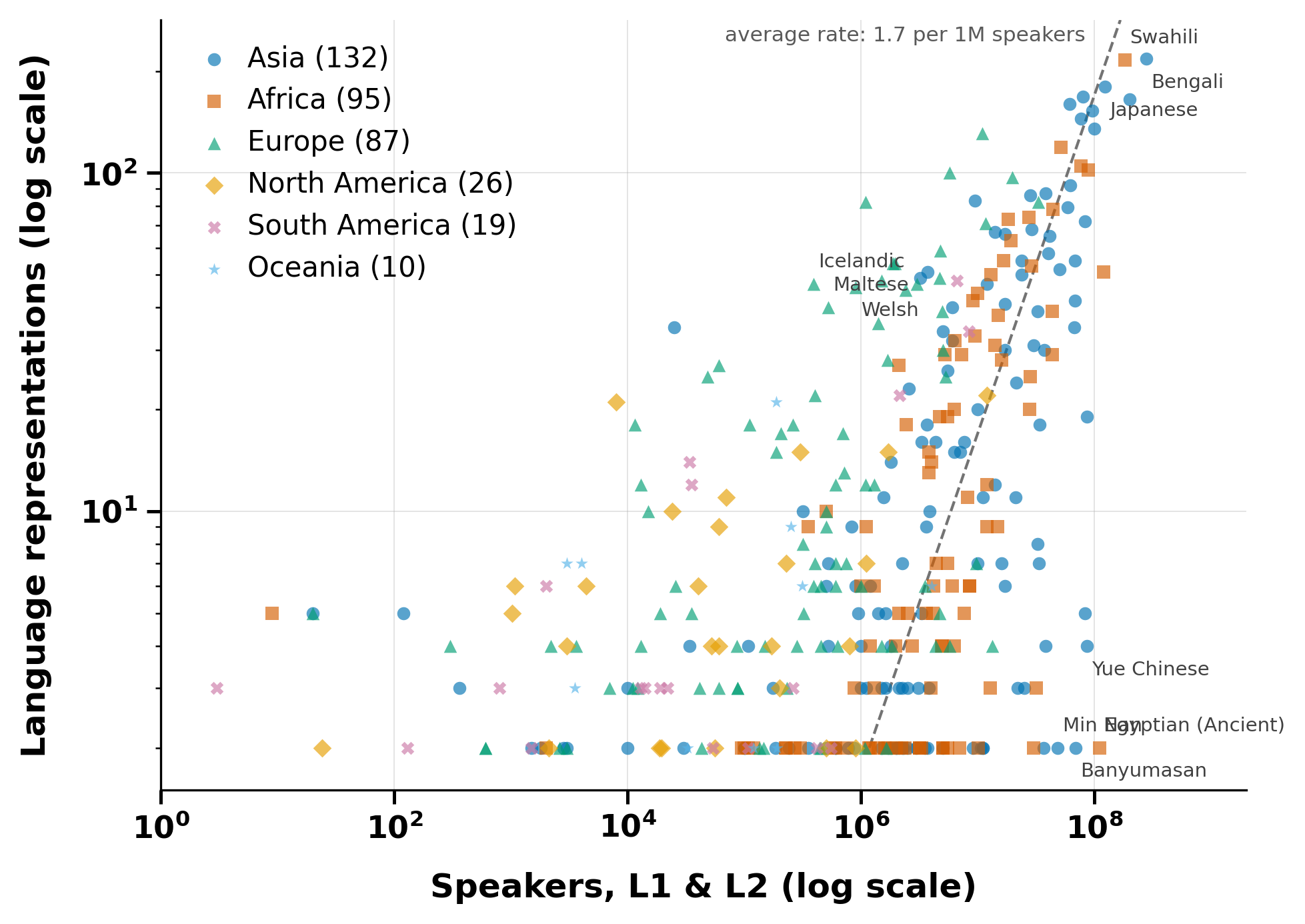}
    \caption{Language representation with respect to the speaker population.}
    \label{fig:Speaker_vs_lang_rep}
\end{figure}

Figure \ref{fig:Speaker_vs_lang_rep} also shows the over- or under-representation of each language extracted from the papers. As shown in the figure, the languages receive a high level of research attention relative to their speaker populations. In contrast, Africa and Asia have the largest share of speakers, but many languages in these regions receive comparatively less attention. Languages spoken in the Americas and Oceania also show relatively high representation compared to their small speaker shares. 

\FloatBarrier
\section{HuggingFace Identification and Resource Collection}
\label{sec:hf-collection}

\paragraph{Languages.} Each target language was specified by its ISO~639-1 and/or ISO~639-3 codes and one or more names or aliases (e.g., \textit{mi}, \textit{mri}, \textit{Maori}, \textit{Māori}). 

\paragraph{Candidate retrieval.} We queried the HuggingFace Hub with the \texttt{huggingface\_hub} Python client, retrieving both models and datasets through four strategies:
(i) structured search on the \texttt{language:<code>} tag for every ISO code of a language,
(ii) free-text search on each language name, alias, and any code of at least four characters (shorter strings such as \textit{am} or \textit{to} match many unrelated repositories),
(iii) free-text search combining each language name with the terms \textit{multilingual}, \textit{afro}, \textit{african languages}, and \textit{masakhane}, and
(iv) direct lookup of 41 models and 25 datasets that are widely used for multilingual NLP but often declare \texttt{multilingual} instead of listing individual languages. The models include mBERT, XLM-R, NLLB-200, mBART-50, M2M-100, SeamlessM4T, mT5, ByT5, MADLAD-400, BLOOM, Whisper, AfroXLMR, and AfriBERTa. The datasets include mC4, CC100, OSCAR, FLORES, FLEURS, Common Voice, WikiANN, Universal Dependencies, Tatoeba, MasakhaNER~2.0, MAFAND, SIB-200, XL-Sum, xP3, XTREME, XNLI, and PAWS-X.
Results were merged and de-duplicated per language by resource type and repository identifier.

\paragraph{Validation.} For each candidate, we downloaded its model or dataset card (YAML metadata and README text) and computed a relevance score from the criteria in Table~\ref{tab:hf-scoring}. Name matching was case-insensitive, restricted to whole words, and limited to terms of at least four characters. Candidates scoring at least 5 were included automatically (high confidence: $\geq 8$, medium: 5--7). Candidates scoring 1--4 were reviewed manually, and candidates scoring 0 were discarded.

\begin{table}[h!]
\centering
\small
\begin{tabular}{lr}
\toprule
\textbf{Criteria} & \textbf{Weight} \\
\midrule
Language code in repository tags & +5 \\
Language code or name in card \texttt{language} field & +5 \\
Language name in card README text$^{*}$ & +5 \\
Language name in tags or card metadata & +3 \\
Language name in repository identifier & +2 \\
Language name in other descriptive fields & +1 \\
\bottomrule
\end{tabular}
\caption{Relevance criteria used to validate Hugging Face candidates. $^{*}$Evaluated only when the card's \texttt{language} field did not match.}
\label{tab:hf-scoring}
\end{table}

\paragraph{Language scope.} Each included repository was labelled monolingual if its card declares exactly one language, multilingual if it declares more than one or the tag \texttt{multilingual}, and unknown if it declares none.

\paragraph{Dating and aggregation.} We define a repository's first availability date as the earlier of its Hub creation timestamp and its earliest commit. This reflects availability on HuggingFace, not necessarily the resource's original release. Counts are computed over language--repository pairs, so a multilingual repository is counted once for each target language it covers; regional totals are sums over languages. For each language, resource type, and year from 2020 to 2025, we count repositories first available in that year, and cumulative counts are running sums from 2020.
\end{document}